\documentclass[10pt,twocolumn,letterpaper]{article}

\usepackage{iccv}
\usepackage{times}
\usepackage{epsfig}
\usepackage{graphicx}
\usepackage{amsmath}
\usepackage{amssymb}
\usepackage{algorithm}
\usepackage{algorithmic}
\usepackage{booktabs}
\usepackage{subcaption}
\usepackage{makecell}

\usepackage[pagebackref=true,breaklinks=true,letterpaper=true,colorlinks,bookmarks=false]{hyperref}

\iccvfinalcopy % *** Uncomment this line for the final submission

\def\iccvPaperID{6530} % *** Enter the ICCV Paper ID here

\ificcvfinal\fi

\begin{document}

%%%%%%%%% TITLE
\title{ShiftNAS: Improving One-shot NAS via Probability Shift}

\author{Mingyang Zhang, Xinyi Yu\thanks{Xinyi Yu and Linlin Ou are corresponding authors.}, Haodong Zhao, Linlin Ou$^{*}$\\
Zhejiang University of Technology\\
Hangzhou, Zhejiang, China\\
{\tt\small yuxy@zjut.edu.cn, linlinou@zjut.edu.cn}
% \and
}

\maketitle
%\ificcvfinal\thispagestyle{empty}\fi

%%%%%%%%% ABSTRACT
\begin{abstract}
   One-shot Neural architecture search (One-shot NAS) has been proposed as a time-efficient approach to obtain optimal subnet architectures and weights under different complexity cases by training only once. However, the subnet performance obtained by weight sharing is often inferior to the performance achieved by retraining. In this paper, we investigate the performance gap and attribute it to the use of uniform sampling, which is a common approach in supernet training. Uniform sampling concentrates training resources on subnets with intermediate computational resources, which are sampled with high probability. However, subnets with different complexity regions require different optimal training strategies for optimal performance.

To address the problem of uniform sampling, we propose ShiftNAS, a method that can adjust the sampling probability based on the complexity of subnets. We achieve this by evaluating the performance variation of subnets with different complexity and designing an architecture generator that can accurately and efficiently provide subnets with the desired complexity. Both the sampling probability and the architecture generator can be trained end-to-end in a gradient-based manner. With ShiftNAS, we can directly obtain the optimal model architecture and parameters for a given computational complexity. We evaluate our approach on multiple visual network models, including convolutional neural networks (CNNs) and vision transformers (ViTs), and demonstrate that ShiftNAS is model-agnostic. Experimental results on ImageNet show that ShiftNAS can improve the performance of one-shot NAS without additional consumption. Source codes are available at \href{https://github.com/bestfleer/ShiftNAS}{GitHub}.
\end{abstract}

%%%%%%%%% BODY TEXT
\begin{figure}[th]
    \centering
    \includegraphics[width=3.25in]{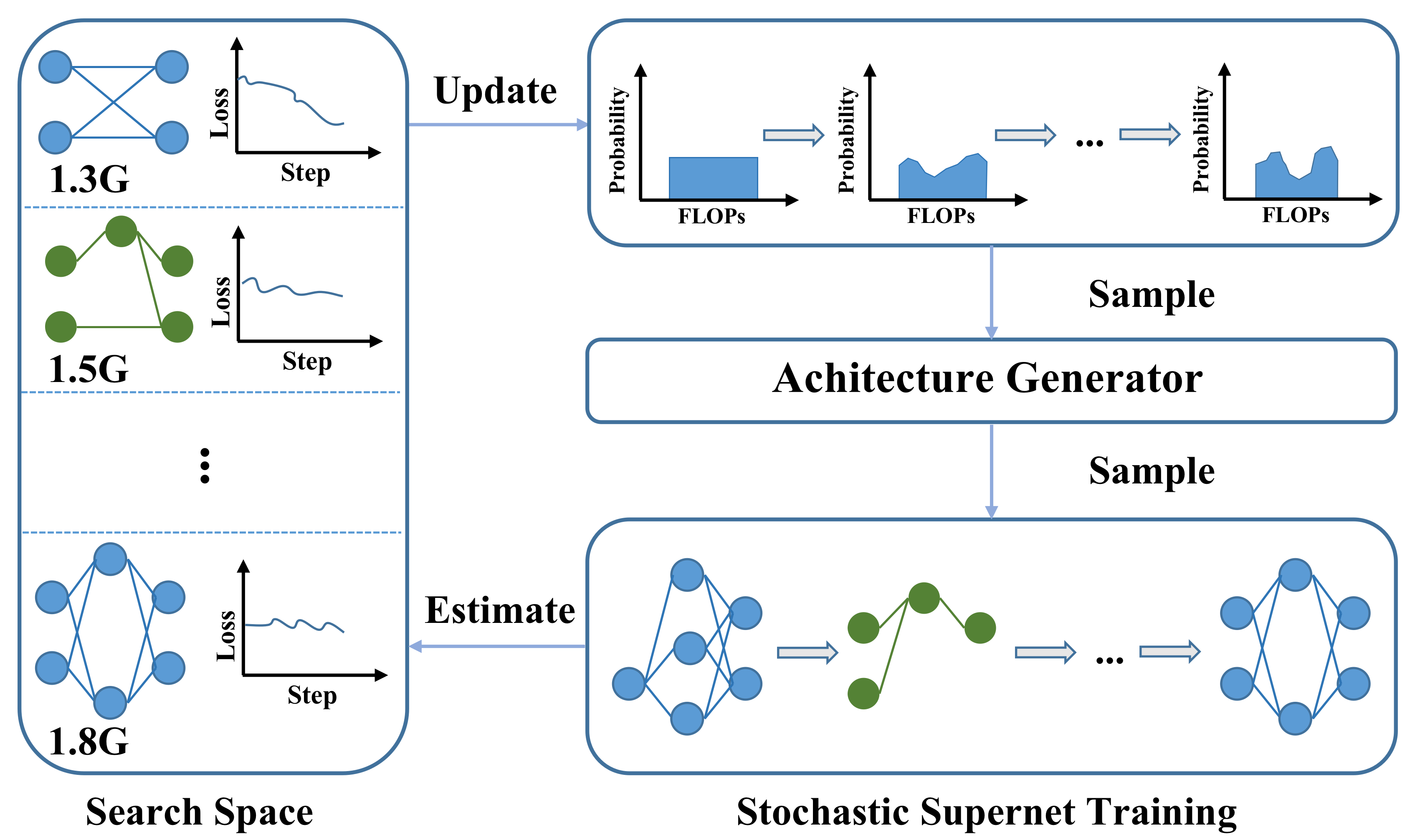}
    \caption{A conceptual overview of our ShiftNAS. Focusing on the computational resource (e.g. FLOPs), ShiftNAS split a search space into several parts where subnets have close computational complexity. In stochastic supernet training, ShiftNAS first samples computational complexity according to probability, then an architecture generator samples the subnet with desired computational complexity. The probability distribution is dynamically updated with stochastic training by estimating the performance variation of each sub-space.}
    \label{fig:overview}
\end{figure}
\section{Introduction}
\label{sec:intro}
Deep neural networks (DNNs) have been widely applied to the field of computer vision with remarkable success \cite{dosovitskiy2020image,he2016deep}. However, the deployment of these vision networks on edge devices still has some limitations, such as the massive model sizes and excessive computation overhead \cite{howard2019searching,howard2017mobilenets,liu2021swin,sandler2018mobilenetv2}. In addition, designing architectures artificially in a trial-and-error manner is a resource-consuming task that requires not only architectural skills but also domain expertise. Consequently, how to acquire optimal architectures that balance latency and accuracy efficiently is of paramount importance.

Recent advancements in neural architecture search (NAS) methods, such as \cite{prenas, guo2020single, wu2019fbnet, zoph2016neural}, have led to significant improvements in the performance of practical applications by automatically searching for optimal architectures within a defined search space. However, traditional NAS methods typically require a substantial computation budget \cite{pham2018efficient, zoph2016neural}. In order to speed up training and reduce the resource consumption of the training process, one-shot NAS methods \cite{liu2018darts, guo2020single} have adopted a two-stage training approach based on weight sharing. Specifically, a supernet is trained in the first stage, and subnets with better performance are searched for in the second stage. In some cases, post-processing methods (e.g., retraining and finetuning) are also necessary in the second stage since the performance of these subnets inherited from the supernet is often inferior to that of models trained from scratch. However, the downside of such methods is that the training consumption increases linearly with the number of architectures, which can be problematic. To address this issue, some one-shot NAS methods \cite{cai2019once, chen2021autoformer, wang2021attentivenas, yu2020bignas} have utilized a weight entanglement training strategy to share the weights in each operation, eliminating the need for additional finetuning or retraining. Furthermore, a high-quality supernet is essential for candidate architectures to inherit weights directly \cite{yu2020bignas}, as well as for accurately ranking candidate architectures \cite{li2020block}.

However, there is still potential for breakthroughs in training a better supernet. In the first stage, previous methods \cite{guo2020single, chen2021autoformer} assume that all candidate architectures are equally important and should be sampled with equal probability during training. However, subnets with different numbers of parameters require different amounts of training resources \cite{cai2021network}. For instance, subnets with 1.0 GFLOPs may converge after 30000 iterations, while subnets with 2.0 GFLOPs may require 50000 iterations. Additionally, only the optimal subnet will be deployed, while others will be ignored at the same computational complexity. Therefore, subnets that occupy more training resources may not be distributed in regions corresponding to resource constraints, resulting in sub-optimal performance of the final deployed model. We find that when all subnets are sampled with equal probability, the resulting computational resource distribution is approximately normal. Consequently, subnets trained under this distribution may appear to be under-fitting or over-fitting in different regions.

To address the challenge of efficiently training a high-performance supernet, we propose a novel method, probability-\textbf{Shift} \textbf{N}eural \textbf{A}rchitecture \textbf{s}earch (\textbf{ShiftNAS}). In ShiftNAS, the sampling probability of subnets is not uniform and can be dynamically adjusted during the training process. The training sufficiency of each subnet is measured by evaluating its performance variation under different computational constraints. The subnets with high-performance variance are identified as undertrained, and their sampling probabilities are increased to tilt training resources dynamically towards them. This enables us to allocate resources more effectively and efficiently to achieve better performance for the subnets that need more training resources.

In spite of having an optimal sampling distribution, efficiently and accurately sampling subnets with a desired computational constraint still poses a challenge. To address this issue, we propose an LSTM-based architecture generator (AG) that can be optimized differentiably with a resource constraint loss function. The AG's output is then processed by Gumbel Softmax \cite{jang2016categorical} to generate a one-hot vector policy for each searched operation. To suit the weight-entangled search space \cite{yu2019autoslim,chen2021autoformer}, we employ a matrix mapping technique that can convert the one-hot vector into a differentiable mask. The mask is multiplied by the operation to obtain a differentiable subnet. The AG and supernet can be jointly trained to learn how to generate the best subnets with desired computational constraints. During evaluation, the AG can generate a corresponding subnet immediately for any given computational constraint. The weights of the searched subnet can be directly inherited from the well-trained supernet, making ShiftNAS free from any additional search or retrain costs.

The overall contribution can be summarized as follows:
\begin{itemize}
    \item A learnable sampling strategy, called probability shift, is proposed to relief the bias of uniform sampling which leads to performance gap between supernet training and subnet deployment.
    \item We propose an LSTM-based AG to precisely and efficiently offer the best subnet with desired resource constraints. AG training can be differentiably trained with supernet under weight-entangled search space.
    \item We achieve state-of-the-art or competitive results on both CNN and ViT models. Therefore, ShiftNAS is a model-agnostic search method.
\end{itemize}

%-------------------------------------------------------------------------
\section{Related Work}
The one-shot NAS method \cite{cai2019once,wang2021attentivenas,yu2020bignas} can automatically search for the optimal architecture in a predefined search space \cite{chen2021autoformer,howard2019searching,liu2018darts}, which can be formulated by maximizing an expected accuracy over the space $A$, i.e.,
\begin{equation}
    \begin{aligned}
        & \min _{\alpha \in A} L_{v a l}\left(w^{*} \mid \alpha\right) \\
        & \text { s.t. } w^{*}=\underset{w_{\alpha}}{\arg \min } L_{\text {train }}(w \mid \alpha) \\
        & complexity(\alpha)<\tau
    \end{aligned}
\end{equation}
where $w$ and $\alpha$ represent the weight and architecture of subnets. $L_{val}$ and $L_{train}$ are loss functions in the validation and train dataset, respectively. The computational complexity of $\alpha$ can be calculated by the complexity function. $\tau$ denotes the complexity threshold for alpha. To solve this optimization problem, a two-stage approach based on weight sharing is usually employed. 

In the first stage, a high-quality supernet is trained by sampling a large number of subnets in the defined search space. Notably, in order to obtain subnets that need to be incorporated into the training phase, the following approaches are commonly used in recent one-shot NAS works \cite{chen2021autoformer,guo2020single}.

First is the uniform sampling method, which is also the most common strategy. The Uniform sample method considers that all architectures in the supernet are equally important, that is, they should be sampled with exactly equal probability. The weight optimization equation in the first stage can be formulated as:
\begin{equation}
    \begin{aligned}
        & \min _{w} E_{\alpha \sim A}\left[L_{\operatorname{train}}(w \mid \alpha)\right] \\
        & \alpha=o^{1} \cup o^{2} \cup \ldots \cup o^{D} \\
        & o^{i} \sim U\left(0, n_{i}\right)
    \end{aligned}
\end{equation}
where $D, n$ denotes the total operation and candidate number in certain operations, respectively. $o^{i}$ denotes the $i$ th selected operation, which follows a uniform distribution.

Since there is a gap between the training process and practical deployment, AttentiveNAS \cite{wang2021attentivenas} proposes the Pareto-aware sampling method. Specifically, models deployed on edge devices are generally around the Pareto frontier, but they are uniformly sampled during training, which leads to the waste of training resources to a certain extent. To bridge the gap between training and deploying, each iteration extracts multiple subnets with equal size and selects the best or worst architecture for training.

Similar to AttentiveNAS, GreedyNAS \cite{you2020greedynas} screen the weak subnets and just sample from the potentially-good subnets instead of all subnets, thus capturing another opportunity to improve the accuracy of target models.

However, these uniformly sampled architectures appear to be an approximately normal distribution of the computational cost. To further reduce the gap between training and deployment, Focusformer \cite{liu2022focusformer} proposes that all the sampled subnets should be based on resource distribution and focuses more on the Pareto frontier architectures through an architecture sampler. In addition, FairNAS \cite{chu2019fairnas} also reinforces the sampling process with a stricter standard of fairness, that is, the parameters of each choice block are updated the same number of times at any stage.

When the first stage is over, this well-trained supernet can be used as the performance estimator of the candidate architecture. And the weight of candidate architectures can be directly inherited from the supernet.

In the second stage, we need to search for the best candidate architecture under different resource constraints in the supernet, which can be formulated as
\begin{equation}
    \begin{gathered}
        \alpha=\underset{\alpha \in A}{\operatorname{argmin}} L_{v a l}(w \mid \alpha) \\
        \text s.t.\ complexity(\alpha)<\tau
    \end{gathered}    
\end{equation}
Since there are so many architectures in the supernet that need to be evaluated, the consumption of computing resources is also unacceptable. Therefore, recent works resort to random search \cite{li2020random}, evolution algorithms \cite{chen2021autoformer,guo2020single,pham2018efficient} or reinforcement learning \cite{zoph2016neural} to find the most promising architecture among all architectures. In the end, these selected architectures need some post-processing operations such as retraining to improve the architecture performance, but this also brings additional computational consumption. 

In order to alleviate the above-mentioned resource consumption problem, some one-shot NAS methods train a high-quality supernet so that candidate architectures can directly inherit supernet weights without retraining, such as Autoformer \cite{chen2021autoformer}, attentiveNAS \cite{wang2021attentivenas}, BigNAS \cite{yu2020bignas}, FocusFormer \cite{liu2022focusformer}.

\section{Nerual Architecture Search with Probability Shift}
In this section, we first discuss the training resources allocation problem caused by uniform sampling in one-shot NAS. To solve the above-mentioned problem, we present ShiftNAS, an end-to-end supernet training framework. In ShiftNAS, a learnable sampling strategy is proposed, which can dynamically adjust the training resources allocation by shifting the sampling probability. Then, an LSTM-based $\mathrm{AG}$ is designed to accurately obtain the expected subnet for each sampling. We show the overview of our method in \textbf{Figure \ref{fig:overview}}.

\subsection{Rethinking the Sampling of One-shot NAS}
In one-shot NAS, an overparameterized supernet $S$ is formed with multiple operations, where each operation contains several choices. The supernet $S$ contains $D$ operations, and the $d$ th operation $o^{d}$ can be selected from $n$ candidates ${o_{1}, \ldots, o_{n}}$, which represent various design choices, such as kernel size and channels in CNN search space \cite{yu2020bignas}, or heads number and MLP ratio in ViT search space \cite{chen2021autoformer}. A subnet $a$ sampled from the supernet can be represented by a tuple of size $D$, i.e., $a=\left(o^{1}, \ldots, o^{D}\right)$.

\begin{figure}[h]
	\centering
	\begin{subfigure}{0.9\linewidth}
		\centering
		\includegraphics[width=0.9\linewidth]{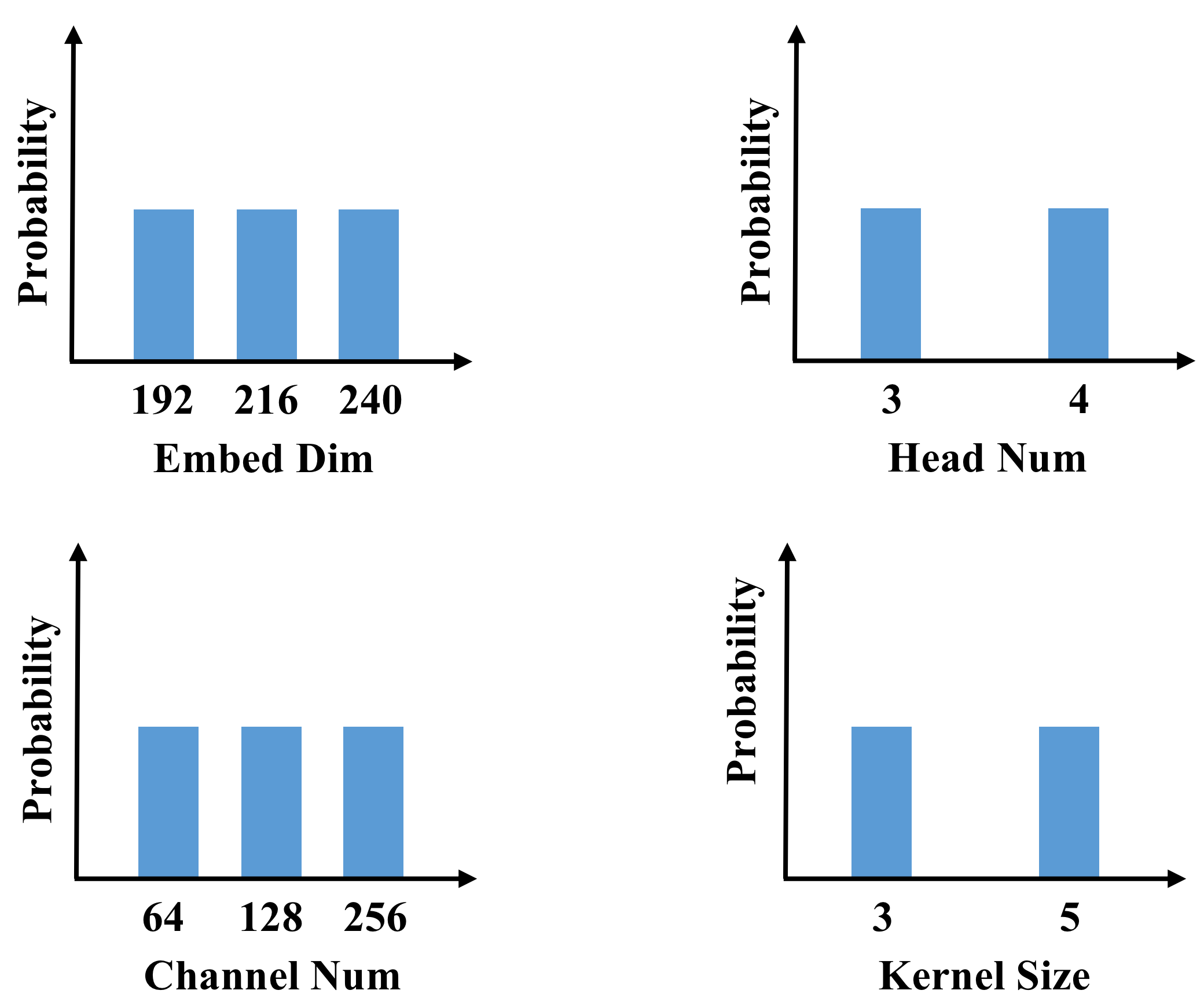}
		\caption{Uniform distribution of each operation}
		\label{fig:uniform-distribution}
	\end{subfigure}
	\begin{subfigure}{0.9\linewidth}
		\centering
		\includegraphics[width=0.9\linewidth]{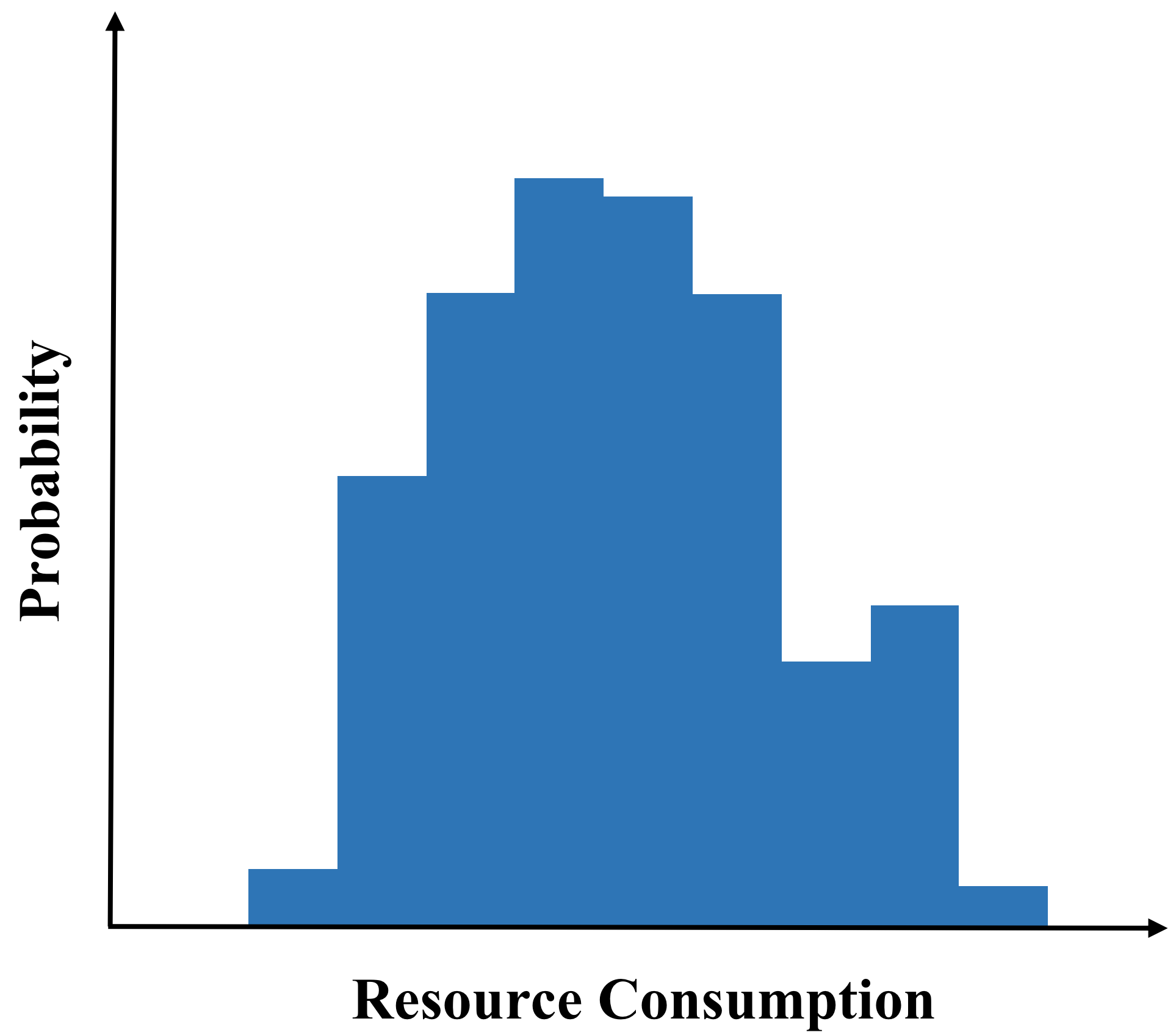}
		\caption{Normal distribution of subnets computational resources}
		\label{fig:norm-distribution}
	\end{subfigure}
 \caption{During supernet training, operations are sampled from a uniform distribution in each iteration. The subnet consists of the sampled operations and its computational resource follows an approximated normal distribution.}
 \label{fig:distribution}
\end{figure}

In previous one-shot NAS methods, the sampling probability of each operation candidate $o_i$ is given by $p\left(o_{i}\right)=\frac{1}{n}$, which assumes that all candidates are equally important. However, in reality, one-shot NAS is more concerned with the computational resources of the subnets. To address this, we introduce the notion of computational resource for each operation candidate $o^{d}$, denoted by $b^{d} \in{b_{1}, \ldots, b_{n}}$. The computational resource of a subnet $a$ randomly sampled from the supernet can then be computed as $B_{a}=\sum_{d=1}^{D} b^{d}$, where $b^{d}$ follows a uniform distribution. 

\newtheorem{remark}{Remark}
\begin{remark}
\label{remark1}
If we assume that the computational resource of each operation $b^{d}$ is independently sampled from a uniform distribution, then the total computational resource $B_{a}$ of a subnet $a$ sampled from the supernet follows an Irwin-Hall distribution. As the number of operations $D$ in the supernet increases, the Irwin-Hall distribution converges to a normal distribution \cite{marengo2017geometric}. 
\end{remark}

Upon analyzing \textbf{Remark \ref{remark1}}, it is evident that the uniform sampling strategy results in subnets being sampled with moderate computational resources, as depicted in \textbf{Figure \ref{fig:distribution}}. This observation implies that subnets with moderate computational resources can be trained effectively, whereas subnets with large or small computational resources cannot be fully trained. Insufficiently-trained subnets lead to inaccurate ranking \cite{li2020block,li2021bossnas} and inherit unreasonable weights from the supernet without retraining \cite{bender2018understanding}.

\begin{figure*}[t]
    \centering
    \includegraphics[width=6.5in]{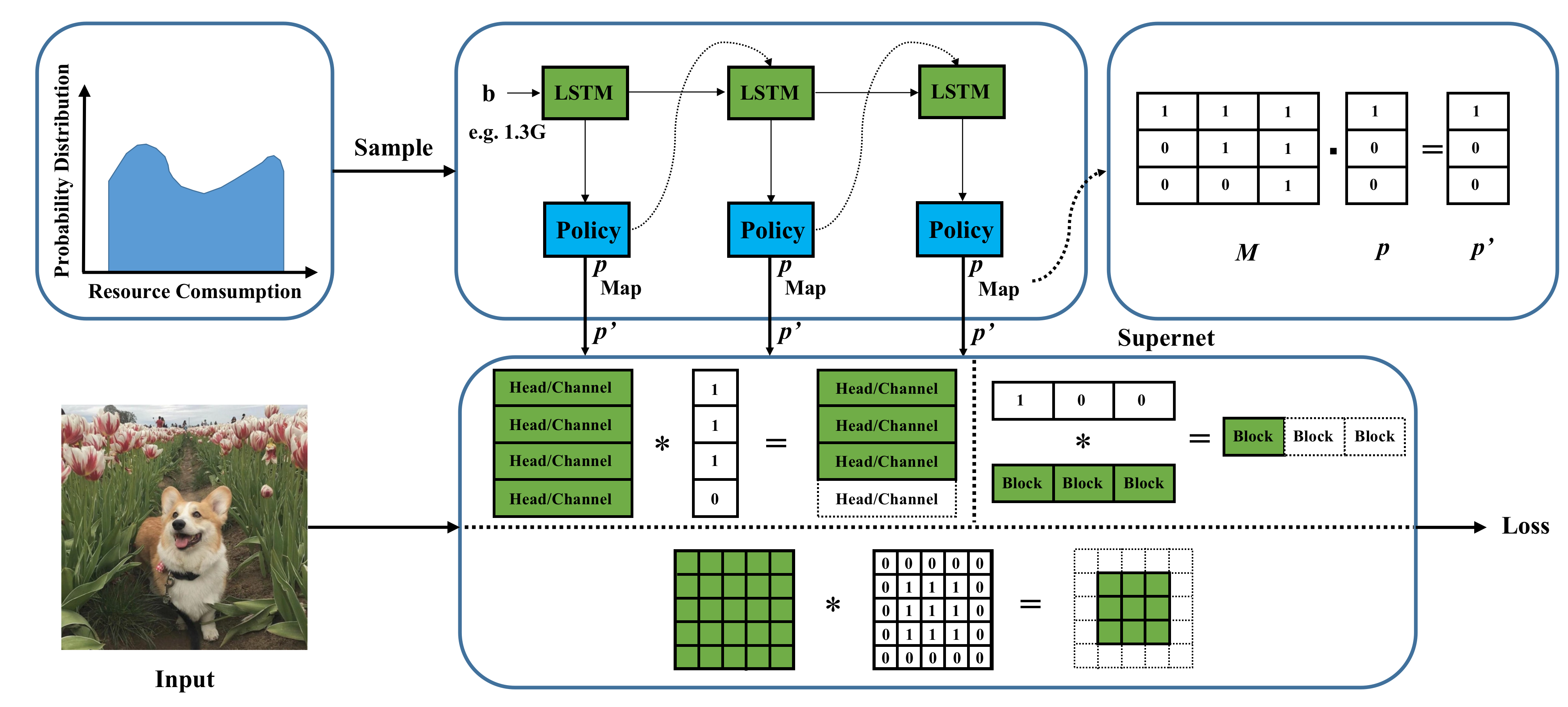}
    \caption{An overview of architecture generator training process. AG is given a sampled resource constraint B and generates several one-hot policies sequence by sequence. Then, one-hot policies will be mapped to corresponding masks to keep differentiable.}
    \label{fig:ag}
\end{figure*}

\subsection{Shifting Probability with Greedy Sample}
One-shot NAS methods rely on the assumption that the supernet can effectively rank the performance of subnets with arbitrary computational resources, and therefore require each subnet to be fully trained with optimal training strategies. However, in practice, subnets with different computational resources may require different training strategies, and the uniform sampling strategy used in previous one-shot NAS methods may lead to insufficient training of subnets in certain computational regions. For example, subnets with extreme computational resources may not be fully trained even when subnets with moderate computational resources are fully trained. To address this issue, we propose a learnable greedy sampling strategy that takes into account the training sufficiency of subnets in different computational regions.

For any subnet, we judge whether the subnet has converged by calculating the following gradient of the subnet.
\begin{equation}
    \label{eq:gradient_w}
    \nabla_{w}=\nabla_{w} L_{val}(w, \alpha \mid b)
\end{equation}
where $L_{val}$ represents the task loss (e.g. cross entropy) in the validation dataset, $w, \alpha$ and $b$ denote the weight, architecture and sampled computational resource of the subnet, respectively. When the subnet with $b$ computational resource converges, $\nabla_{w}$ will tend to zero. In other words, insufficiency-trained subnets have a large gradient and should be sampled more times. Therefore, we can greedily adjust the sampling distribution of $B$ by optimizing the computational resource distribution at each training step
\begin{equation}
    \label{eq:optimize}
    \underset{B}{\arg \max } E_{b \sim B}\left[\nabla_{w} L_{v a l}(w, \alpha \mid b)\right]
\end{equation}
To optimize $B$ end to end based on gradient-based methods, we discrete computational resources into several parts, e.g. $[1.3 G, 1.4 G, \ldots, 2.0 G]$ FLOPs, and uniformly initialize its probability distribution. In each training step, one computational resource can be sampled by Gumbel Softmax $G$. Formally, \textbf{Eq. \ref{eq:optimize}} can be rewritten as
\begin{equation}
    \label{eq:optimizer_with_g}
    \underset{B}{\arg \max } \nabla_{w} L_{val}(w, \alpha \mid G(B))    
\end{equation}
In this way, the gradient of $B$ in $t$th update can be obtained by applying chain rule to \textbf{Eq. \ref{eq:optimizer_with_g}}
\begin{equation}
    \label{eq:2-order_gradient}
    \nabla_{B_t}=\nabla_{w_t, B_t}^{2} L_{val}(w_t, \alpha \mid G(B_t)) \nabla_{B_t} G(B_t)
\end{equation}
where $t$ denotes the update times of $B$. $w_t$ represents the supernet weight when $B$ has been updated $t$ times. However, evaluating the gradient of $B_t$ requires an expensive matrix-vector product in its second term. Inspired by darts \cite{liu2018darts}, the complexity can be substantially reduced using the finite difference approximation. We use two training steps to approximate the gradient of $w$, therefore, \textbf{Eq. \ref{eq:2-order_gradient}} can be rewritten as
\begin{equation}
    \label{eq:gradient_b}
    \begin{array}{r}
        \nabla_{B_t}=\nabla_{B_t} L_{val}\left(w_{t}, \alpha \mid G(B_t)\right) \nabla_{B_t} G(B_t) \\
        -\nabla_{B_t} L_{val}\left(w_{t-1}, \alpha \mid G(B_t)\right) \nabla_{B_t} G(B_t)
    \end{array}
\end{equation}
In the implementation, the distribution $B$ is only updated every few iterations, so there is no excessive computational overhead. The iterative procedure is outlined in \textbf{Appendix}.

\subsection{Generating Architectures with Arbitray Computational Resources}
The probability shift is able to sample subnets according to their corresponding training sufficiency. However, a crucial question emerges: given a certain computational resource $b$, how can we rapidly sample a subnet that satisfies the desired resource constraint? A straightforward approach would be to keep sampling subnets until one is found that meets the resource constraint. Nevertheless, as illustrated in \textbf{Figure \ref{fig:distribution}}, subnets with large or small computational resources have an extremely low probability of being sampled, which makes sampling these subnets computationally inefficient.

To address this issue, we propose an architecture generator (AG) that can provide the corresponding subnet architecture according to any resource constraint. The AG is designed to stably generate a subnet architecture that satisfies the desired resource constraint. Inspired by previous works \cite{zoph2016neural}, which formulate the NAS problem as a sequence prediction problem, we also use an LSTM network to generate each operation sequence by sequence. To update the sampling probability $B$, we use differentiable neural architecture search (DNAS) \cite{hu2020dsnas,liu2018darts} to jointly train the AG and the supernet. DNAS is preferred over other search methods, such as RL-based or evolutionary-based methods, as it converges faster.

This section presents an overview of the AG (Architecture Generator) training process with clarity and academic rigor. Figure \ref{fig:ag} provides a visual representation of the AG training process. During the training of AG and supernet, the sample distribution $B$ remains static and is not updated. To enable end-to-end updates of AG, the policy of each operation is generated using Gumbel Softmax with a one-hot vector, such as $[1,0,0]$.

It is important to note that the focus of this paper is on weight entanglement. This means that each operation shares weights for their common parts, and weights with small indices are always activated. As a result, policies cannot be directly involved in forward and backward processes. To overcome this issue, a matrix map trick is employed. More specifically, given an one-hot policy $p \in 1 \times n$ where $n$ denotes the candidate number in this operation, the mapped policy $p^{\prime}$ can be computed using the following
\begin{equation}
    \label{eq:map}
    p^{\prime}=M \cdot p^T=[m^T_0, m^T_1, \ldots, m^T_n]\cdot\ p^T
\end{equation}
where $M$ consists of several masks and $m^T_i$ represents the mask of $i$ th candidate. We show the matrix map example in \textbf{Figure \ref{fig:ag}}. Assume a ViT block has three heads and the heads number can be selected in $[1,2,3]$, their weight-entangle masks can be designed as $m^{0}=$ $[1,0,0], m^{1}=[1,1,0]$ and $m^{2}=[1,1,1]$, respectively. With \textbf{Eq. \ref{eq:map}}, the one-hot policy $p$ participates in the forward of the supernet while the gradient of $\mathrm{AG}$ is automatically calculated by the chain rule.

\begin{table*}[t]
\begin{center}
\begin{tabular}{c|c|c|c|c|c}
\toprule[1pt]
Model & Parameters(M) & FLOPs(G) & Cost(GPU Days) & Top-1 Acc.(\%) & Top-5 Acc.(\%) \\
\toprule[1pt]
DeiT-tiny\cite{touvron2021training}&5.7&1.2&24&72.2&91.1\\
AutoFormer-tiny\cite{chen2021autoformer}&5.7&1.3&30&74.7&92.3\\
FocusFormer-tiny\cite{liu2022focusformer}&6.2&1.4&26&75.1&93.1\\
\textbf{ShiftFormer-T(Ours)}&5.8&\textbf{1.3}&\textbf{24}&\textbf{76.0}&\textbf{93.1}\\
\toprule[1pt]
DeiT-S\cite{touvron2021training}&22.1&4.7&30&79.9&95.0\\
T2T-ViT-14\cite{yuan2021tokens}&21.5&6.1&32&81.7&-\\
ViT-S/16\cite{dosovitskiy2020image}&22.9&5.1&-&78.8&-\\
BoTNet-S1-59\cite{srinivas2021bottleneck}&33.5&7.3&-&81.7&95.8\\
AutoFormer-small\cite{chen2021autoformer}&22.9&5.1&35&81.4&95.6\\
FocusFormer-small\cite{liu2022focusformer}&23.7&5.0&32&81.6&95.6\\
\textbf{ShiftFormer-S(Ours)}&23.6&\textbf{5.0}&\textbf{30}&\textbf{82.2}&\textbf{95.8}\\
\toprule[1pt]
DeiT-B\cite{touvron2021training}&86.6&17.6&43&81.8&95.6\\
ViT-B/16\cite{dosovitskiy2020image}&86.6&17.6&-&79.7&-\\
AutoFormer-base\cite{chen2021autoformer}&52.8&11.0&43&81.4&95.7\\
FocusFormer-base\cite{liu2022focusformer}&52.8&11.0&41&81.9&95.6\\
\textbf{ShiftFormer-B(Ours)}&52.8&\textbf{11.0}&\textbf{40}&\textbf{82.8}&\textbf{96.1}\\
\toprule[1pt]
MobileNetV2 0.75$\times$\cite{sandler2018mobilenetv2}&2.6&0.21&18&69.8&-\\
MobileNetV3 1.0$\times$\cite{howard2019searching}&5.4&0.22&3791&75.2&-\\
DS-MBNet-M\cite{9842348}&-&0.33&24&73.2&-\\
BigNAS-S\cite{yu2020bignas}&4.5&0.24&112&76.5&-\\
Once-For-All\cite{cai2019once}&4.4&0.23&105&76.4&-\\
GreedyNAS-C\cite{you2020greedynas}&4.7&0.28&32&76.2&92.5\\
\textbf{ShiftCNN-S(Ours)}&4.5&\textbf{0.24}&\textbf{32}&\textbf{77.2}&\textbf{93.1}\\
\toprule[1pt]
MobileNetV2 1.3$\times$\cite{sandler2018mobilenetv2}&5.3&0.50&21&72.8&-\\
MobileNetV3 1.25$\times$\cite{howard2019searching}&8.1&0.35&3791&76.7&-\\
EfficientNet-B0\cite{tan2019efficientnet}&5.4&0.39&3791&77.1&93.3\\
DS-MBNet-S\cite{9842348}&-&0.57&24&74.8&-\\
BigNAS-M\cite{yu2020bignas}&5.5&0.42&112&78.9&-\\
GreedyNAS-A\cite{you2020greedynas}&6.5&0.37&40&77.1&93.3\\
\textbf{ShiftCNN-B(Ours)}&5.6&0.42&\textbf{32}&\textbf{79.6}&\textbf{93.6}\\
\toprule[1pt]
\end{tabular}
\caption{ShiftNAS models performance on ImageNet with comparisons to other models. We group the models according to their FLOPs. We use ShiftFormer and ShiftCNN to denote the models searched by ShiftNAS. \textbf{Cost} represents the total GPU days including training, searching and retraining.}
\label{tab:main_result}
\end{center}
\end{table*}
Since our goal is to make $\mathrm{AG}$ learn how to accurately generate a subnet architecture with arbitrary resource constraints, the objective function of $\mathrm{AG}$ is designed to reduce task loss $L_{task}$ while minimizing the gap $L_{R C}$ between the resource constraint of the sampled subnet and the target. Therefore, the joint loss $L$ is given by
\begin{equation}
    \begin{aligned}
        & L=L_{\text {task }}+\lambda L_{R C} \\
        & L_{R C}=\left(\sum_{i=1}^{D} \sum_{j=1}^{n} b_{j}^{i} p_{j}^{\prime i}-C\right)^{2}
    \end{aligned}
    \label{eq:optim_ag}
\end{equation}
where $b_{j}^{i}$ and $p_{j}^{\prime i}$ represents the computational resource and policy of $j$ th candidate in $i$ th operation, $C$ indicates the target resource constraint sampled from sampling distribution B. $\lambda$ is the coefficient of $L_{R C}$. In the implementation, we jointly optimize the AG and the supernet on the train dataset with \textbf{Eq. \ref{eq:optim_ag}}. After few epochs, AG can generate the corresponding subnets with given resource constraint.
\begin{figure*}[t]
    \centering
    \includegraphics[width=7in]{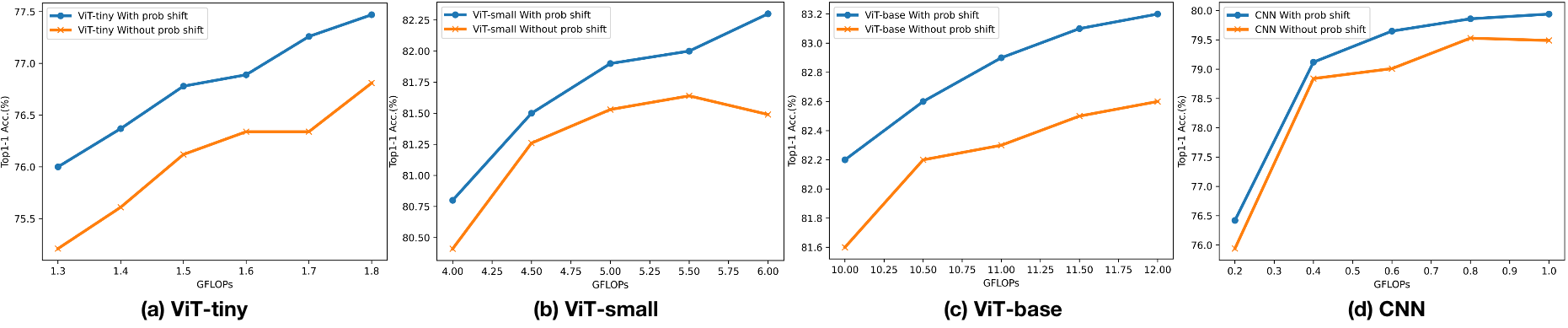}
    \caption{The effectiveness of probability shift. FLOPs/Accuracy tradeoffs of ShiftNAS with and without probability shift on search space (a) ViT-tiny; (b) ViT-small; (c) ViT-base; (d) CNN.}
    \label{fig:acc-flops-curve}
\end{figure*}
\section{Experiment}
\subsection{Experimental Setup}
\textbf{Dataset and evaluation metrics.} We conduct main experiments on ImageNet \cite{krizhevsky2012imagenet} which contains about 1.2M images for training and $50 \mathrm{~K}$ images for validation. We use the Top-1 accuracy and the number of giga floating-point operations(GFLOPs) to measure the performance and efficiency of networks.

\textbf{Implementation details.} ShiftNAS is a model-agnostic search method. Therefore, we search the models with various resource constraints on both CNN and ViT search space. We split the computational resource range into $K$ parts with $0.1$ GFLOPs step. The sampling distribution will be represented by a $1 \times K$ learnable vector. We randomly split $50 \mathrm{~K}$ images from ImageNet and use Adam optimizer with 1e-3 learning rate to update the sampling distribution vector. The AG consists of an LSTM cell with 64 hidden numbers and 4/10 fully connected layers for ViT/CNN supernet and will be updated jointly with the supernet. In the first 50 training epochs, we only optimize the supernet weight and the AG since the AG cannot generate the corresponding subnets at the beginning. After 50 epochs, the sampling distribution vector is updated per 100 iteration. We discuss more hyperparameter settings in the Appendix, such as the split step of the computational resource and the updated frequency of the sampling distribution vector. The training strategies for both CNN and ViT are given in Appendix. All the models are trained on 8 Nvidia Tesla A100 GPUs.
\begin{table}[t]
\begin{center}
\begin{tabular}{c|c|c|c}
\toprule[1pt]
Model & Inherit(\%) & Finetune(\%) & Retrain(\%) \\
\toprule[1pt]
ShiftFormer-T & 76.0 & 75.9(-0.1) & 76.1(+0.1) \\
ShiftFormer-S & 82.0 & 82.0(0.0) & 82.0(0.0) \\
ShiftFormer-B & 82.4 & 82.3(-0.1) & 82.6(+0.2) \\
ShiftCNN-S & 77.2 & 77.0(-0.2) & 77.3(+0.1) \\
ShiftCNN-B & 79.6 & 79.5(-0.1) & 79.6(0.0) \\
\toprule[1pt]
\end{tabular}
\caption{Comparison of subnets Top-1 accuracy with inherited, finetuned and retrained weights.}
\label{tab:retrain}
\vspace{-2em}
\end{center}
\end{table}

For CNN search space, we follow BigNAS \cite{yu2020bignas} where the search space contains kernel size, channel number, depth and input resolution. The supernet will be split into 7 stages. Each stage has multiple choices of the block number and the first block of each stage has no residual path. The policy of kernel size will be generated block-wise, where the small kernel size is center cropped from the large kernel size. The policies of channel number and depth will be predicted stage-wise, where the lower-index channels and blocks are preferentially kept.  

\begin{figure*}[th]
    \centering
    \includegraphics[width=7in]{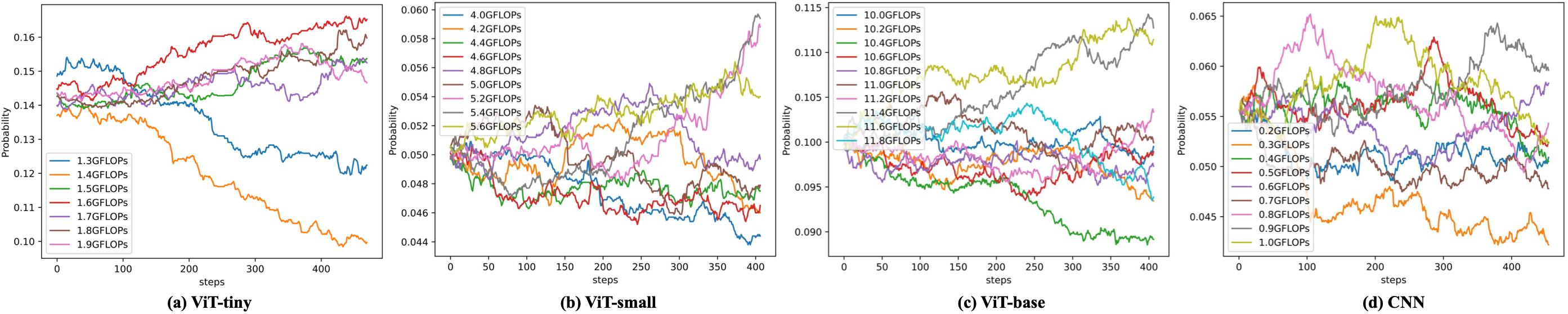}
    \caption{The sampling probability with different FLOPs groups on search space (a) ViT-tiny; (b) ViT-small; (c) ViT-base; (d) CNN.}
    \label{fig:strategy}
\end{figure*}

For ViT search space, we follow AutoFormer \cite{chen2021autoformer} which searches embed dim, head number, MLP ratio and depth with three different scale settings. The policies of head number and MLP ratio will be given block-wise. As CNN, the lower-index heads and neurons are preferentially kept. 

After training, we directly sample the candidate with the max probability given by the AG. It is noted that the obtained model inherits weights from the trained supernet without retraining or fine-tuning. \textbf{Therefore, no search or retraining cost exists in ShiftNAS.}
\begin{figure}[h]
    \centering
    \includegraphics[width=2.8in]{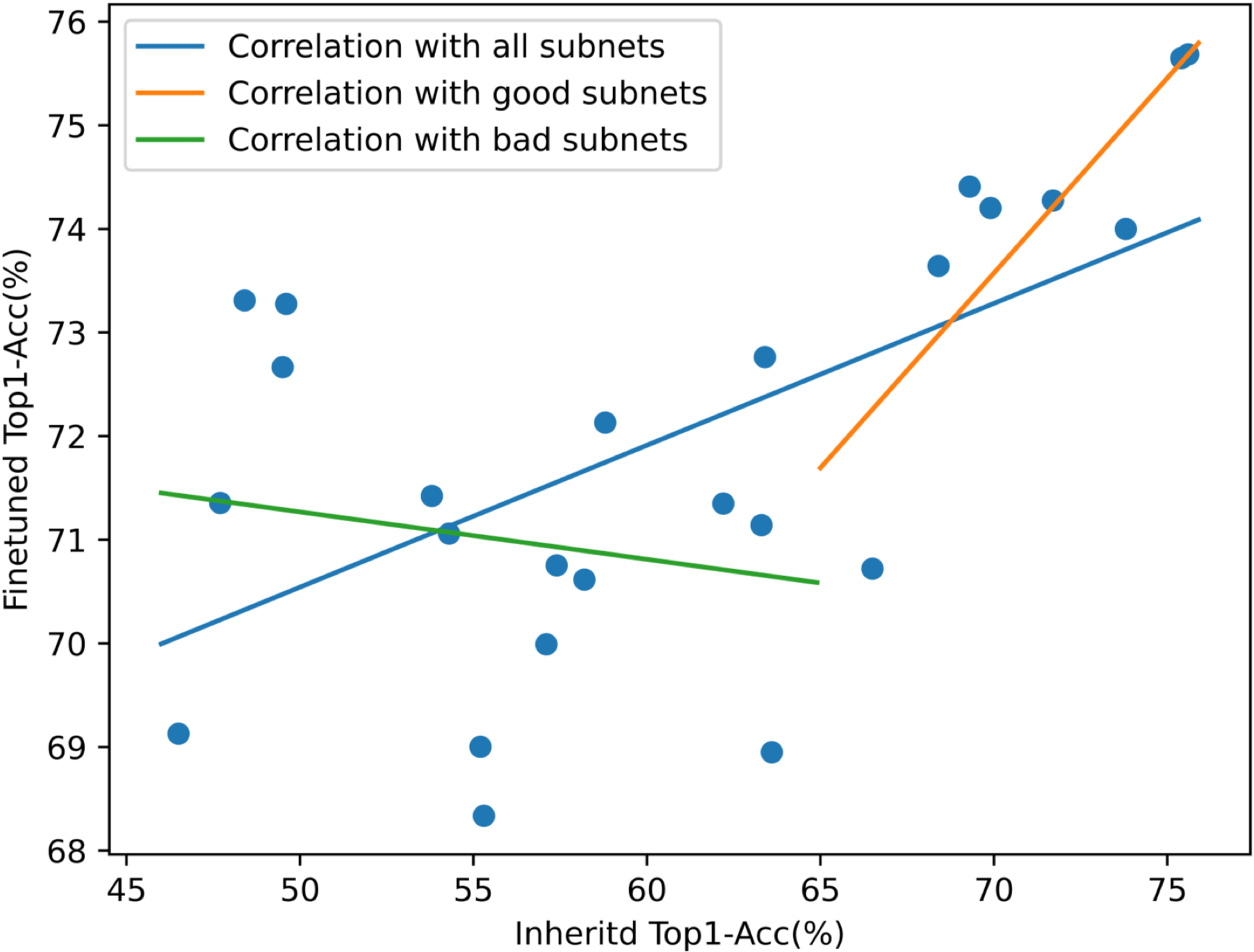}
    \caption{The ranking correlation of architecture generator.}
    \label{fig:ranking}
\end{figure}
\subsection{Main Results}
We search ViT and CNN architectures on ImageNet with different FLOPs constraints. We compare searched models with multiple ViT and CNN models on model performance (Top-1 Acc., Top-5 Acc.) and efficiency (FLOPs, parameters). It is seen from \textbf{Table \ref{tab:main_result}} that our ShiftFormer models surpass the recent manual and autoML-based transformer models under diverse model sizes. It is worth noting that ShiftFormer-tiny achieves a top-1 Acc. of $76.0$ with only $1.3$ GFLOPs, being $1.3 \%$ and $0.9 \%$ better than AutoFormer-tiny and FocusFormer-T which are also autoML-based methods, respectively. Compared to CNN models, ShiftCNN models also achieve better accuracy than all compared networks under similar FLOPs restrictions. For example, ShiftCNN-S and ShiftCNN-B achieve $0.7 \%$ and $0.7 \%$ higher top-1 accuracy than BigNAS-S and BigNAS-M, respectively. Specifically, ShiftCNN and BigNAS models are searched from the same search space. BigNAS models spend more than $2 \times$ GPU time than ShiftCNN models since the sandwich rule is used to train BigNAS supernet, which demonstrates the efficiency and outstanding performance of ShiftNAS. For fairness, we do not compare our models with others obtained from better search spaces \cite{li2021bossnas,wang2021attentivenas} in \textbf{Table \ref{tab:main_result}}. The experiments on AttentiveNAS search space can be found in the \textbf{Appendix}.

\subsection{Ablation Study}
\textbf{The effectiveness of probability shift.} To demonstrate the effectiveness of the proposed sampling probability shift, we trained a ShiftNAS supernet and its baseline counterpart under ViT-tiny, ViT-small, ViT-base, and CNN search spaces. In the baseline supernet training, we uniformly sampled FLOPs in each step and did not update FLOPs probability. The training settings for all supernets are provided in the Appendix.
As illustrated in Figure \ref{fig:acc-flops-curve}, the supernet trained with probability shift is capable of covering a wide range of accuracy-FLOPs curves and outperforms the baseline supernet trained without probability shift by a significant margin, demonstrating the effectiveness of the proposed sampling probability shift mechanism. 

\textbf{The ranking correlation of architecture generator.} The experiment is based on ShiftFormer-tiny supernet. We measure the ranking correlation by randomly sampling 25 subnets from the ShiftFormer-tiny supernet. For fast evaluation, The sampled subnets are inherited weights from the supernet and are further finetuned to reach the optimal weights. As for the model ranking test, we evaluate sampled subnets in both supernets to get their inherited performance. As shown in \textbf{Figure \ref{fig:ranking}}, the inherited performance is highly correlated with finetuned performance in good subnets (Inherited Top1-Acc $>65 \%)$. Despite the correlation with bad subnets is not obvious, we focus more attention on good subnets since bad subnets are not deployed in practice.

\textbf{Subnet performance without retraining.} We further study the performance of ShiftNAS models when models are independently fine-tuned or trained from scratch. Following Autoformer, we finetune or retrain the selected subnet with 40 epochs or 300 epochs on ImageNet. \textbf{Table \ref{tab:retrain}} shows that finetuning cannot improve or even harm the performance of subnets inherited from supernet. Besides, retraining can slightly improve the subnet performance but cost extra training resources which prevents the implementation of NAS.

\textbf{The optimal sampling strategy.} ShiftNAS learns the optimal sampling probability for each subnet with different FLOPs. To figure out what sampling strategy is the best, we visualize the various sampling probability in different search spaces (ViT-tiny, ViT-small, ViT-base and CNN) at different training steps. As can be seen in \textbf{Figure \ref{fig:strategy}}, our method learns to sample more subnets with large computational complexity. For example, the sampling probabilities of $1.6$ to $1.9$ GFLOPs subnets are clearly higher than $1.3$ and $1.4$ GFLOPs subnets in the ViT-tiny search space. This is intuitive since the large subnet contains more parameters and needs more computational resources to update.
\section{Conclusion}
In this paper, we presented ShiftNAS, a one-shot supernet training framework that can complexity-wise adjust the sampling probability. To automatically adjust the sampling probability, we proposed probability shift that can be learned according to the subnets training sufficiency. To fast obtain the subnets with desired complexity, AG is designed that can be jointly trained with the supernet in an end-to-end manner. Experiments on ImageNet showed that our method achieves SOTA results on ViT/CNN search spaces, and covers a wide range of efficiency/accuracy trade-offs without any extra retraining. 

\textbf{Limitation.} ShiftNAS is designed for efficient search on weight-entanglement search spaces, so it cannot work on darts-like search spaces \cite{liu2018darts,siems2020bench}.

{\bf Acknowledgements:} This work was supported by the National Natural Science Foundation of China (62203392) and Natural Science Foundation of Zhejiang pvovince (LY21F030018).

{\small
\bibliographystyle{ieee_fullname}
\bibliography{egbib}

\begin{thebibliography}{10}\itemsep=-1pt

\bibitem{bender2018understanding}
Gabriel Bender, Pieter-Jan Kindermans, Barret Zoph, Vijay Vasudevan, and Quoc
  Le.
\newblock Understanding and simplifying one-shot architecture search.
\newblock In {\em International conference on machine learning}, pages
  550--559. PMLR, 2018.

\bibitem{cai2021network}
Han Cai, Chuang Gan, Ji Lin, and Song Han.
\newblock Network augmentation for tiny deep learning.
\newblock {\em arXiv preprint arXiv:2110.08890}, 2021.

\bibitem{cai2019once}
Han Cai, Chuang Gan, Tianzhe Wang, Zhekai Zhang, and Song Han.
\newblock Once-for-all: Train one network and specialize it for efficient
  deployment.
\newblock In {\em International Conference on Learning Representations}, 2019.

\bibitem{chen2021autoformer}
Minghao Chen, Houwen Peng, Jianlong Fu, and Haibin Ling.
\newblock Autoformer: Searching transformers for visual recognition.
\newblock In {\em Proceedings of the IEEE/CVF international conference on
  computer vision}, pages 12270--12280, 2021.

\bibitem{chu2019fairnas}
Xiangxiang Chu, Bo Zhang, Ruijun Xu, and Jixiang Li.
\newblock Fairnas: Rethinking evaluation fairness of weight sharing neural
  architecture search.
\newblock {\em arXiv preprint arXiv:1907.01845}, 2019.

\bibitem{dosovitskiy2020image}
Alexey Dosovitskiy, Lucas Beyer, Alexander Kolesnikov, Dirk Weissenborn,
  Xiaohua Zhai, Thomas Unterthiner, Mostafa Dehghani, Matthias Minderer, Georg
  Heigold, Sylvain Gelly, et~al.
\newblock An image is worth 16x16 words: Transformers for image recognition at
  scale.
\newblock {\em arXiv preprint arXiv:2010.11929}, 2020.

\bibitem{guo2020single}
Zichao Guo, Xiangyu Zhang, Haoyuan Mu, Wen Heng, Zechun Liu, Yichen Wei, and
  Jian Sun.
\newblock Single path one-shot neural architecture search with uniform
  sampling.
\newblock In {\em European Conference on Computer Vision}, pages 544--560.
  Springer, 2020.

\bibitem{he2016deep}
Kaiming He, Xiangyu Zhang, Shaoqing Ren, and Jian Sun.
\newblock Deep residual learning for image recognition.
\newblock In {\em Proceedings of the IEEE conference on computer vision and
  pattern recognition}, pages 770--778, 2016.

\bibitem{howard2019searching}
Andrew Howard, Mark Sandler, Grace Chu, Liang-Chieh Chen, Bo Chen, Mingxing
  Tan, Weijun Wang, Yukun Zhu, Ruoming Pang, Vijay Vasudevan, et~al.
\newblock Searching for mobilenetv3.
\newblock In {\em Proceedings of the IEEE/CVF international conference on
  computer vision}, pages 1314--1324, 2019.

\bibitem{howard2017mobilenets}
Andrew~G Howard, Menglong Zhu, Bo Chen, Dmitry Kalenichenko, Weijun Wang,
  Tobias Weyand, Marco Andreetto, and Hartwig Adam.
\newblock Mobilenets: Efficient convolutional neural networks for mobile vision
  applications.
\newblock {\em arXiv preprint arXiv:1704.04861}, 2017.

\bibitem{hu2020dsnas}
Shoukang Hu, Sirui Xie, Hehui Zheng, Chunxiao Liu, Jianping Shi, Xunying Liu,
  and Dahua Lin.
\newblock Dsnas: Direct neural architecture search without parameter
  retraining.
\newblock In {\em Proceedings of the IEEE/CVF Conference on Computer Vision and
  Pattern Recognition}, pages 12084--12092, 2020.

\bibitem{jang2016categorical}
Eric Jang, Shixiang Gu, and Ben Poole.
\newblock Categorical reparameterization with gumbel-softmax.
\newblock 2016.

\bibitem{krizhevsky2012imagenet}
Alex Krizhevsky, Ilya Sutskever, and Geoffrey~E Hinton.
\newblock Imagenet classification with deep convolutional neural networks.
\newblock {\em Advances in neural information processing systems},
  25:1097--1105, 2012.

\bibitem{li2020block}
Changlin Li, Jiefeng Peng, Liuchun Yuan, Guangrun Wang, Xiaodan Liang, Liang
  Lin, and Xiaojun Chang.
\newblock Block-wisely supervised neural architecture search with knowledge
  distillation.
\newblock In {\em Proceedings of the IEEE/CVF Conference on Computer Vision and
  Pattern Recognition}, pages 1989--1998, 2020.

\bibitem{li2021bossnas}
Changlin Li, Tao Tang, Guangrun Wang, Jiefeng Peng, Bing Wang, Xiaodan Liang,
  and Xiaojun Chang.
\newblock Bossnas: Exploring hybrid cnn-transformers with block-wisely
  self-supervised neural architecture search.
\newblock In {\em Proceedings of the IEEE/CVF International Conference on
  Computer Vision}, pages 12281--12291, 2021.

\bibitem{9842348}
Changlin Li, Guangrun Wang, Bing Wang, Xiaodan Liang, Zhihui Li, and Xiaojun
  Chang.
\newblock Ds-net++: Dynamic weight slicing for efficient inference in cnns and
  vision transformers.
\newblock {\em IEEE Transactions on Pattern Analysis and Machine Intelligence},
  pages 1--16, 2022.

\bibitem{li2020random}
Liam Li and Ameet Talwalkar.
\newblock Random search and reproducibility for neural architecture search.
\newblock In {\em Uncertainty in artificial intelligence}, pages 367--377.
  PMLR, 2020.

\bibitem{liu2018darts}
Hanxiao Liu, Karen Simonyan, and Yiming Yang.
\newblock Darts: Differentiable architecture search.
\newblock In {\em International Conference on Learning Representations}, 2018.

\bibitem{liu2022focusformer}
Jing Liu, Jianfei Cai, and Bohan Zhuang.
\newblock Focusformer: Focusing on what we need via architecture sampler.
\newblock {\em arXiv preprint arXiv:2208.10861}, 2022.

\bibitem{liu2021swin}
Ze Liu, Yutong Lin, Yue Cao, Han Hu, Yixuan Wei, Zheng Zhang, Stephen Lin, and
  Baining Guo.
\newblock Swin transformer: Hierarchical vision transformer using shifted
  windows.
\newblock In {\em Proceedings of the IEEE/CVF international conference on
  computer vision}, pages 10012--10022, 2021.

\bibitem{marengo2017geometric}
James~E Marengo, David~L Farnsworth, and Lucas Stefanic.
\newblock A geometric derivation of the irwin-hall distribution.
\newblock {\em International Journal of Mathematics and Mathematical Sciences},
  2017, 2017.

\bibitem{prenas}
Yameng Peng, Andy Song, Vic Ciesielski, Haytham~M. Fayek, and Xiaojun Chang.
\newblock Pre-nas: Evolutionary neural architecture search with predictor.
\newblock {\em IEEE Transactions on Evolutionary Computation}, 27(1):26--36,
  2023.

\bibitem{pham2018efficient}
Hieu Pham, Melody Guan, Barret Zoph, Quoc Le, and Jeff Dean.
\newblock Efficient neural architecture search via parameters sharing.
\newblock In {\em International conference on machine learning}, pages
  4095--4104. PMLR, 2018.

\bibitem{sandler2018mobilenetv2}
Mark Sandler, Andrew Howard, Menglong Zhu, Andrey Zhmoginov, and Liang-Chieh
  Chen.
\newblock Mobilenetv2: Inverted residuals and linear bottlenecks.
\newblock In {\em Proceedings of the IEEE conference on computer vision and
  pattern recognition}, pages 4510--4520, 2018.

\bibitem{siems2020bench}
Julien Siems, Lucas Zimmer, Arber Zela, Jovita Lukasik, Margret Keuper, and
  Frank Hutter.
\newblock Nas-bench-301 and the case for surrogate benchmarks for neural
  architecture search.
\newblock {\em arXiv preprint arXiv:2008.09777}, 2020.

\bibitem{srinivas2021bottleneck}
Aravind Srinivas, Tsung-Yi Lin, Niki Parmar, Jonathon Shlens, Pieter Abbeel,
  and Ashish Vaswani.
\newblock Bottleneck transformers for visual recognition.
\newblock In {\em Proceedings of the IEEE/CVF conference on computer vision and
  pattern recognition}, pages 16519--16529, 2021.

\bibitem{tan2019efficientnet}
Mingxing Tan and Quoc Le.
\newblock Efficientnet: Rethinking model scaling for convolutional neural
  networks.
\newblock In {\em International Conference on Machine Learning}, pages
  6105--6114. PMLR, 2019.

\bibitem{touvron2021training}
Hugo Touvron, Matthieu Cord, Matthijs Douze, Francisco Massa, Alexandre
  Sablayrolles, and Herv{\'e} J{\'e}gou.
\newblock Training data-efficient image transformers \& distillation through
  attention.
\newblock In {\em International conference on machine learning}, pages
  10347--10357. PMLR, 2021.

\bibitem{wang2021attentivenas}
Dilin Wang, Meng Li, Chengyue Gong, and Vikas Chandra.
\newblock Attentivenas: Improving neural architecture search via attentive
  sampling.
\newblock In {\em Proceedings of the IEEE/CVF conference on computer vision and
  pattern recognition}, pages 6418--6427, 2021.

\bibitem{wu2019fbnet}
Bichen Wu, Xiaoliang Dai, Peizhao Zhang, Yanghan Wang, Fei Sun, Yiming Wu,
  Yuandong Tian, Peter Vajda, Yangqing Jia, and Kurt Keutzer.
\newblock Fbnet: Hardware-aware efficient convnet design via differentiable
  neural architecture search.
\newblock In {\em Proceedings of the IEEE/CVF Conference on Computer Vision and
  Pattern Recognition}, pages 10734--10742, 2019.

\bibitem{you2020greedynas}
Shan You, Tao Huang, Mingmin Yang, Fei Wang, Chen Qian, and Changshui Zhang.
\newblock Greedynas: Towards fast one-shot nas with greedy supernet.
\newblock In {\em Proceedings of the IEEE/CVF Conference on Computer Vision and
  Pattern Recognition}, pages 1999--2008, 2020.

\bibitem{yu2019autoslim}
Jiahui Yu and Thomas Huang.
\newblock Autoslim: Towards one-shot architecture search for channel numbers.
\newblock {\em arXiv preprint arXiv:1903.11728}, 2019.

\bibitem{yu2020bignas}
Jiahui Yu, Pengchong Jin, Hanxiao Liu, Gabriel Bender, Pieter-Jan Kindermans,
  Mingxing Tan, Thomas Huang, Xiaodan Song, Ruoming Pang, and Quoc Le.
\newblock Bignas: Scaling up neural architecture search with big single-stage
  models.
\newblock In {\em European Conference on Computer Vision}, pages 702--717.
  Springer, 2020.

\bibitem{yu2018slimmable}
Jiahui Yu, Linjie Yang, Ning Xu, Jianchao Yang, and Thomas Huang.
\newblock Slimmable neural networks.
\newblock In {\em International Conference on Learning Representations}, 2018.

\bibitem{yuan2021tokens}
Li Yuan, Yunpeng Chen, Tao Wang, Weihao Yu, Yujun Shi, Zi-Hang Jiang,
  Francis~EH Tay, Jiashi Feng, and Shuicheng Yan.
\newblock Tokens-to-token vit: Training vision transformers from scratch on
  imagenet.
\newblock In {\em Proceedings of the IEEE/CVF international conference on
  computer vision}, pages 558--567, 2021.

\bibitem{zhang2022segvit}
Bowen Zhang, Zhi Tian, Quan Tang, Xiangxiang Chu, Xiaolin Wei, Chunhua Shen,
  and Yifan Liu.
\newblock Segvit: Semantic segmentation with plain vision transformers.
\newblock {\em arXiv preprint arXiv:2210.05844}, 2022.

\bibitem{zoph2016neural}
Barret Zoph and Quoc~V Le.
\newblock Neural architecture search with reinforcement learning.
\newblock {\em arXiv preprint arXiv:1611.01578}, 2016.

\end{thebibliography}
}

\appendix
\clearpage
\section{Training Settings}
The training settings for ViT and CNN models are following \textbf{Table \ref{tab:training_setting}}. 
\begin{table*}[ht]
\begin{center}
\begin{tabular}{c|c|c|c|c|c|c}
\toprule[1pt]
Model&Epochs&Batch size&Learning rate& Weight decay&Optimizer&Augmentation\\
\toprule[1pt]
CNN&500&1024&5e-1&1e-5&SGD&CropFlip+AutoAugment\\
\toprule[1pt]
ViT&500&1024&1e-3&5e-2&AdamW&\makecell{CropFlip+RandAugment\\ +Cutmix+Mixup+random erasing}\\
\toprule[1pt]
\end{tabular}
\caption{Experimental configurations.}
\label{tab:training_setting}
\vspace{-2em}
\end{center}
\end{table*}

\begin{algorithm}[h]
\caption{The Pytorch-style algorithm of supernet training.}
\label{alg:supernet}
\begin{algorithmic} 
\REQUIRE Supernet architecture $S$ and weight $w$; Architecture generator $AG$; Sampling distribution $\mathbf{B}$; training dataset $(X, Y)$; criterion $\mathrm{C}$; update frequency $q$; Optimizer for $w$ and $\alpha$, $optimizer$.
\ENSURE Trained supernet weight $\mathbf{w}$.
\STATE Initialize frequency count and checkpoint weight, $count$ = 0 and $w_{t-1}$=$w$;
\WHILE{not convergence}
\STATE Clear gradients, $optimizer.zero\_grad()$;
\STATE Sample a subnet computational resource$\mathbf{b}$, $\mathbf{b}\sim{~}\mathbf{B}$;
\STATE Sample a subnet architecture $a$, $a$ = $AG(\mathbf{b}|\alpha)$ ;
\STATE Sample mini-batch of data, $(x, y) \leftarrow(X, Y)$;
\STATE Compute loss, $loss$=$\mathrm{C}$($S(x|w, a), y$);
\STATE Compute $\nabla_w$ $\nabla_\alpha$, $loss$.backward();
\STATE Update $w$ and $\alpha$, $optimizer.step()$;
\STATE $count$+=1;
\IF{$count$==$q$}
\STATE Update $w_t$=$w$ and $count$=$0$;
\STATE Do \textbf{Algorithm \ref{alg:shiftnas}} with $w_t$ and $w_{t-1}$;
\STATE Update $w_{t-1}$=$w_{t}$;
\ENDIF
\ENDWHILE
\end{algorithmic}
\end{algorithm}
\begin{algorithm}[h]
\caption{The Pytorch-style algorithm of distribution update.}
\label{alg:shiftnas}
\begin{algorithmic} 
\REQUIRE Supernet architecture $S$; Checkpoint and current weights $w_{t-1}$, $w_t$; Architecture generator $AG$; Sampling distribution $\mathbf{B}$; validation dataset $(X', Y')$; criterion $\mathrm{C}$; Optimizer for $B$, $optimizer\_B$.
\ENSURE Updated distribution $\mathbf{B}$.
\STATE Clear gradients, $optimizer\_B.zero\_grad()$;
\STATE Sample a subnet computational resource, $\mathbf{b}\sim{~}\mathbf{B}$;
\STATE Sample a subnet architecture $a$, $a$ = $AG(\mathbf{b}|\alpha)$ ;
\STATE Sample mini-batch of data, $(x', y') \leftarrow(X', Y')$;
\STATE Compute $loss_{t-1}$, $loss_{t-1}$ = $\mathrm{C}$($S(x'| w_{t-1},a)$,$y'$);
\STATE Compute and save $\nabla_{B_{t-1}}$, $loss_{t-1}.backward()$;
\STATE Clear gradients, $optimizer\_B.zero\_grad()$;
\STATE Compute $loss_{t}$, $loss_{t}$ = $\mathrm{C}$($S(x' | w_{t}, a)$,$y'$);
\STATE Compute and save $\nabla_{B_{t}}$, $loss_{t}.backward()$;
\STATE Compute $\nabla_{B}$=$\nabla_{B_{t-1}}$-$\nabla_{B_{t}}$;
\STATE Update $B$, $optimizer\_B.step()$;
\end{algorithmic}
\end{algorithm}

\section{Training Algorithm}
We provide a detailed account of the supernet training process in \textbf{Algorithm \ref{alg:supernet}}. Unlike the uniform sampling approach, we propose a dynamic distribution $B$ to sample subnets in each iteration. The sampling distribution $B$ is updated every $q$ iterations as shown in \textbf{Algorithm \ref{alg:shiftnas}}. We calculate $\nabla_{B}$ by performing two forward and backward passes on the current and former supernet weight $w_t$ and $w_{t-1}$. It should be noted that only a batch of data is used in each $B$ update, which keeps the time overhead at an acceptable level.

\section{More Ablation Study}
\textbf{Correlation between the gradient and the training sufficiency.} In ShiftNAS, we utilize the gradient of $\nabla_{B}$=$\nabla_{B_{t-1}}$-$\nabla_{B_{t}}$ to quantify the training sufficiency of subnets, and there is a curiosity about the relationship between the two. To address this, we conducted an experiment on ViT-tiny space to investigate their correlation. Specifically, we followed the steps: \textbf{1)} We trained a supernet with few epochs using a uniform sampling strategy. \textbf{2)} We randomly selected 30 subnets from the supernet and calculated their scores $\nabla_{B}$=$\nabla_{B_{t-1}}$-$\nabla_{B_{t}}$ based on the validation dataset. \textbf{3)} These subnets were independently finetuned for one epoch. \textbf{4)} After finetuning, the loss variations of the sampled subnets on the validation dataset were recorded.

The Kendall's tau values between the scores $\nabla_{B}$ and the loss variations are presented in \textbf{Table \ref{tab:kendall tau}}. Our results demonstrate a strong correlation between the gradient and the training sufficiency after training the supernet for 60 epochs.
\begin{table}[!t]
\begin{center}
\begin{tabular}[c]{c|c|c|c|c}
\toprule[2pt]
Epoch&30&60&90&120\\
\toprule[1pt]
Kendall's tau&0.24&0.63&0.75&0.72\\
\bottomrule[2pt]
\end{tabular}
\caption{The Kendall’s tau values in different training stages.}
\label{tab:kendall tau}
\end{center}
\vspace{-2em}
\end{table}

\begin{figure}[h]
    \centering
    \includegraphics[width=3.7in]{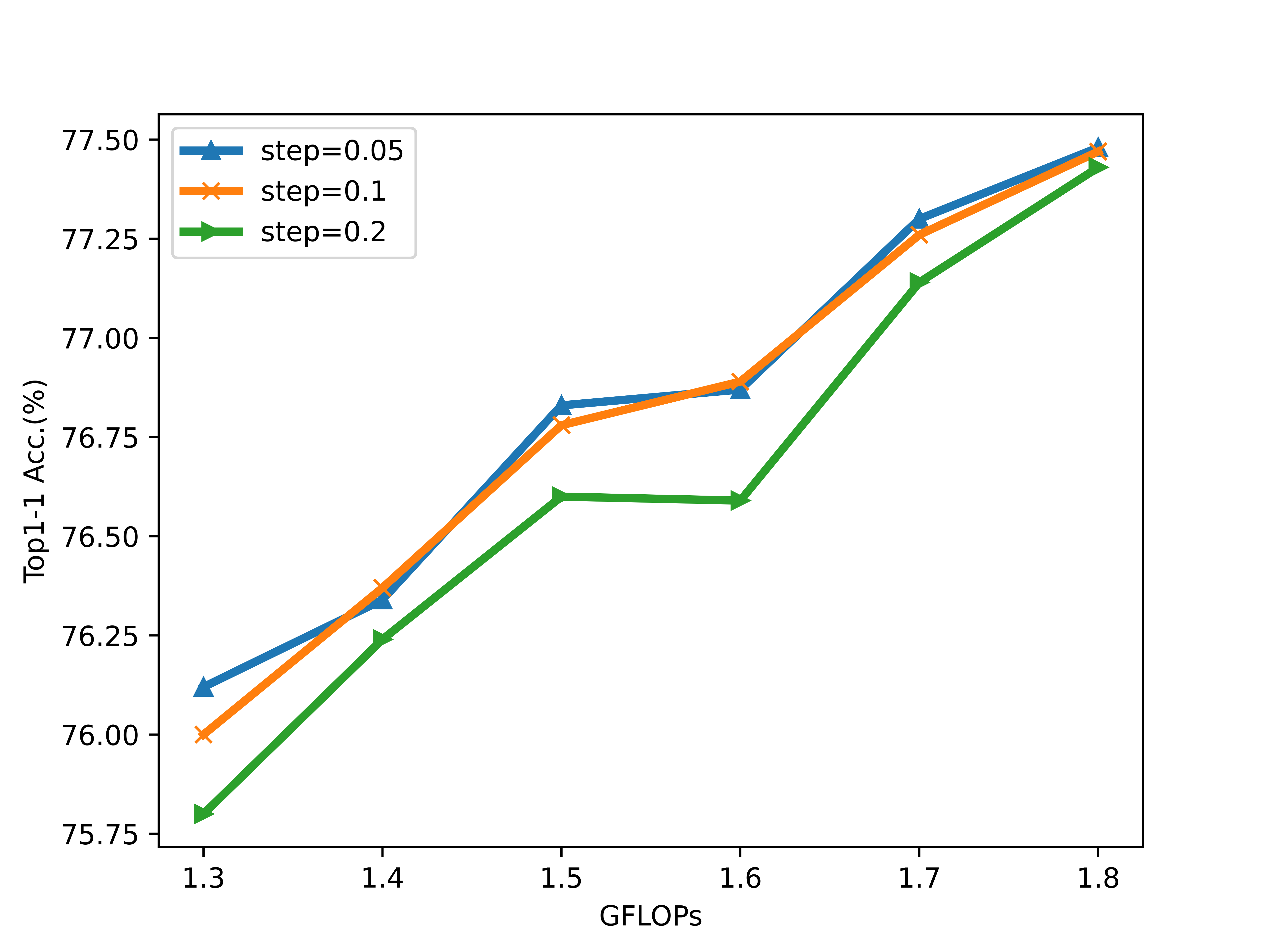}
    \caption{FLOPs/Accuracy tradeoffs of ShiftNAS with different steps}
    \label{fig:steps}
\vspace{-2em}
\end{figure}

\textbf{Split steps of search space.} In ShiftNAS, the search space is divided into several parts based on computational complexity, e.g., FLOPs. The effect of steps on model performance is discussed here, using experiments carried out on ViT-tiny space where each supernet is trained under the same training setting. The search space is split from 1.3 GFLOPs to 1.9 GFLOPs with 0.2, 0.1, and 0.05 GFLOPs steps, respectively. As shown in \textbf{Figure \ref{fig:steps}}, it can be observed that the 0.1 step obtains the best performance in most cases. Empirically, a larger step leads to a smaller search space since the AG only needs to search the optimal subnets along the steps. Therefore, these subnets sampled from a smaller search space can be trained more sufficiently, which is also mentioned in \cite{liu2022focusformer}. However, a large step means that we cannot obtain a fine-grained optimal subnet. Therefore, 0.1 steps are chosen for ShiftNAS to balance performance and deployment.

\begin{figure}[h]
    \centering
    \includegraphics[width=3.7in]{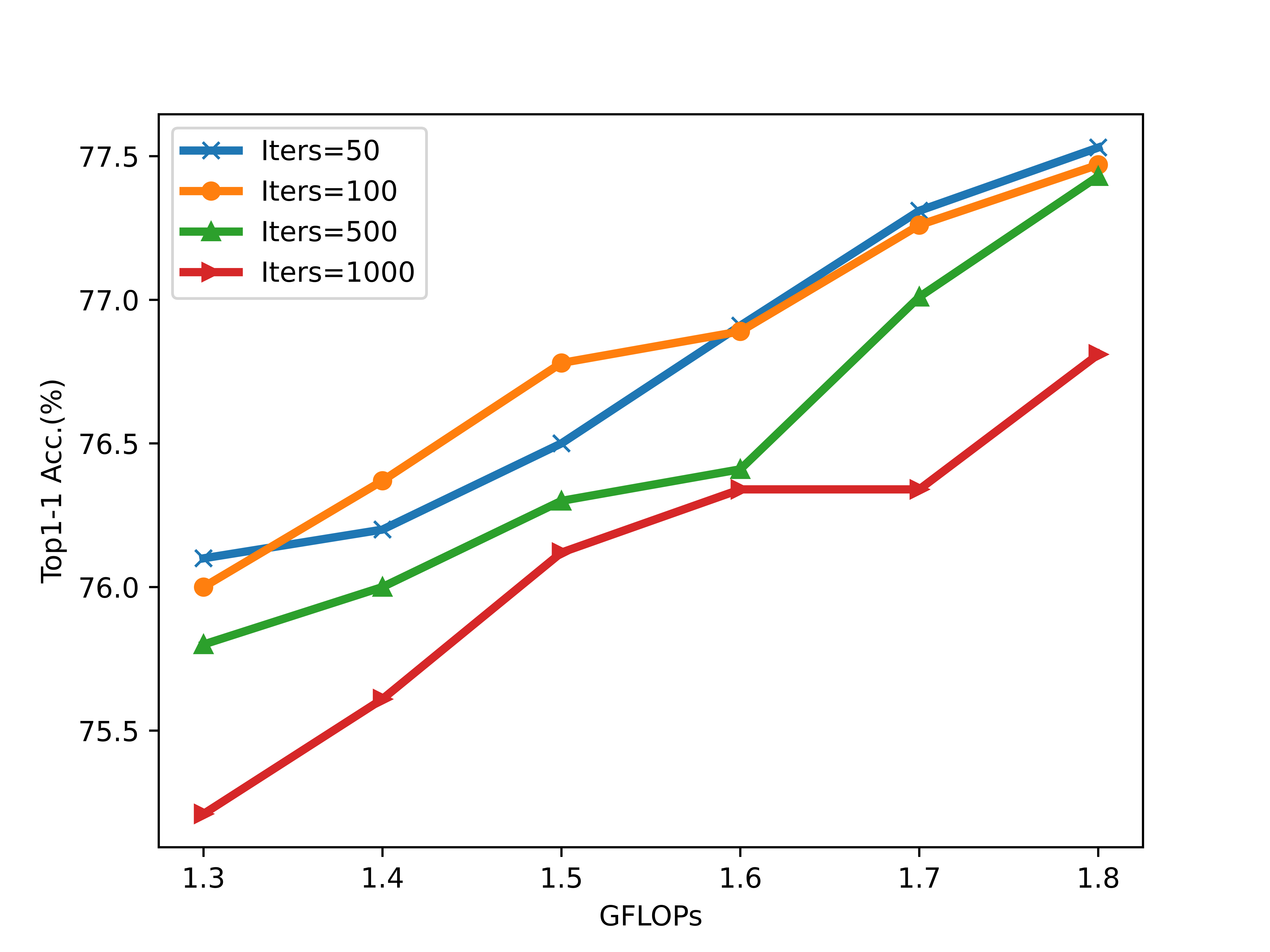}
    \caption{FLOPs/Accuracy tradeoffs of ShiftNAS with different update frequency}
    \label{fig:freq}
\end{figure}

\textbf{Update frequency of the sampling distribution vector.} To investigate the effect of different update frequency, we conducted experiments by setting the update frequency $q$ as 50, 100, 500, and 1000 iterations. The results, shown in \textbf{Figure \ref{fig:freq}}, indicate that the subnet performance decreases as the update frequency decreases, for different computational constraints. This phenomenon suggests that the optimal sampling distribution varies under different training stages, and frequent updates can help to better adapt to the changing training dynamics.

\begin{figure*}[t]
    \centering
    \includegraphics[width=7in]{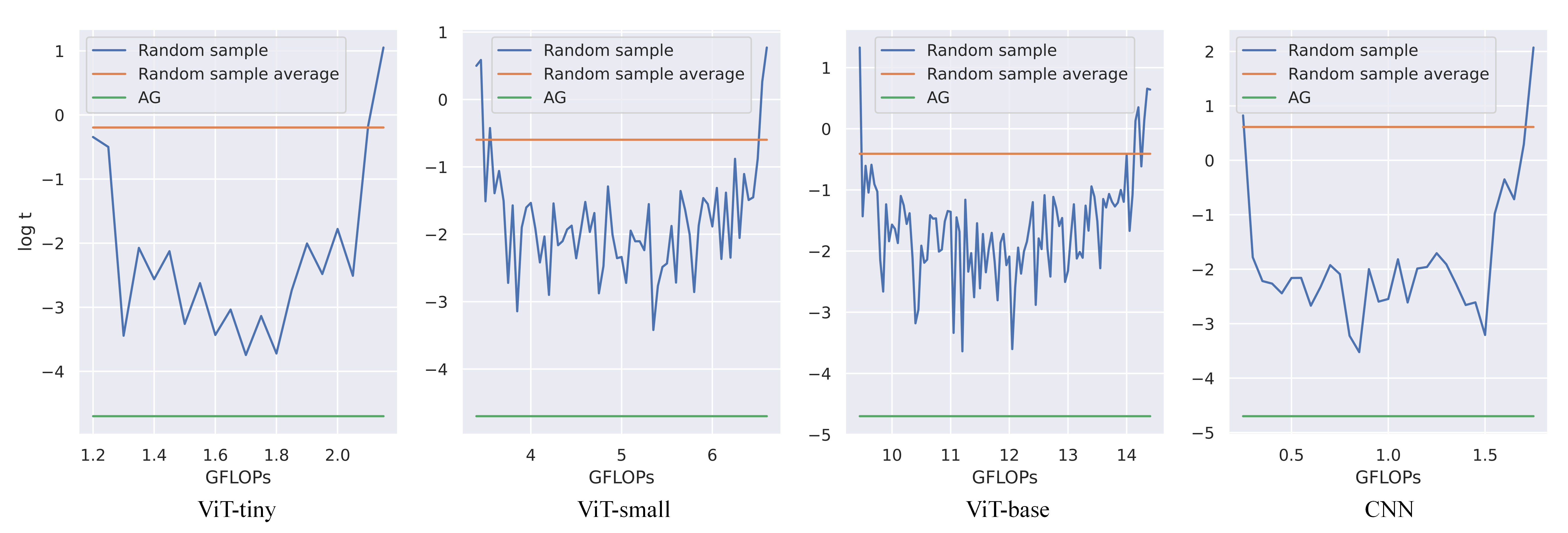}
    \caption{\label{fig:sampling time}Sampling time comparison. Y label is $log t$.}
\end{figure*}

\textbf{The efficiency of architecture generator.} To validate the efficiency of the architecture generator (AG), we compared the time required for AG and without AG when sampling subnets of different FLOPs. Without AG, we randomly sample subnets until it finds one that meets the computational constraint. The experimental results, as shown in Figure \ref{fig:sampling time}, demonstrate that AG can directly infer architectures of any computational complexity, while random sample takes hundreds or thousands of times longer. For example, during searching ViT-tiny on ImageNet-1k, an additional 52 hours ($1250iters \times 500epochs \times 0.4s$) is required.

\section{Comparisons under AttentiveNAS search space}
AttentiveNAS \cite{wang2021attentivenas} has introduced a method to dynamically sample subnets during supernet training. However, this method employs a more comprehensive search space than ours, as illustrated in \textbf{Figure 9 in the AttentiveNAS Appendix}. To ensure a fair comparison, we have trained the supernet with \textbf{500 epochs on AttentiveNAS search space}. The comparative results are presented in \textbf{Table \ref{tab:compare_with_atten}}. Notably, our ShiftCNN models can outperform the AttentiveNAS models under comparable FLOPs constraints. Additionally, it is worth mentioning that ShiftNAS consumes fewer training epochs than AttentiveNAS due to the utilization of the sandwich rule \cite{yu2018slimmable} in training the supernet.
\begin{table}[!t]
\begin{center}
\begin{tabular}[c]{c|c|c|c}
\toprule[2pt]
method&Acc.@1&FLOPs(M)&Training epochs\\
\toprule[1pt]
\textbf{ShiftCNN-S}&\textbf{78.7}&\textbf{259}&\textbf{500}\\
AttentiveNAS-A1&78.4&279&360$\times$4=1440\\
\toprule[1pt]
\textbf{ShiftCNN-B}&\textbf{80.4}&\textbf{453}&\textbf{500}\\
AttentiveNAS-A4&79.8&444&360$\times$4=1440\\
\bottomrule[2pt]
\end{tabular}
\caption{Comparison of the effectiveness with AttentiveNAS.}
\label{tab:compare_with_atten}
\vspace{-2em}
\end{center}
\end{table}

\section{Transfer for Segmentation Tasks}
To assess the transferability of our proposed approach to other computer vision tasks, we have conducted experiments on segmentation using the ADE20k dataset. In this regard, we have employed SegViT \cite{zhang2022segvit} as our framework and have replaced its backbone with ShiftFormer-B. For fair comparison, the baseline backbone is ViT-Base \cite{dosovitskiy2020image}, pre-trained on ImageNet1k. The experimental results are reported in \textbf{Table \ref{tab:compare_on_seg}}.
\begin{table}[t]
\begin{center}
\begin{tabular}[c]{c|c|c}
\toprule[2pt]
Backbone&Acc.@1&FLOPs(G)\\
\toprule[1pt]
ViT-Base&48.2&120.9\\
ShiftFormer-B&\textbf{49.8}&\textbf{90.4}\\
\bottomrule[2pt]
\end{tabular}
\caption{Comparison of the effectiveness on the segmentaion task.}
\label{tab:compare_on_seg}
\vspace{-2em}
\end{center}
\end{table}

\section{Visualization of the Searched Architectures}
\begin{figure*}[h]
    \centering
    \includegraphics[width=7.0in]{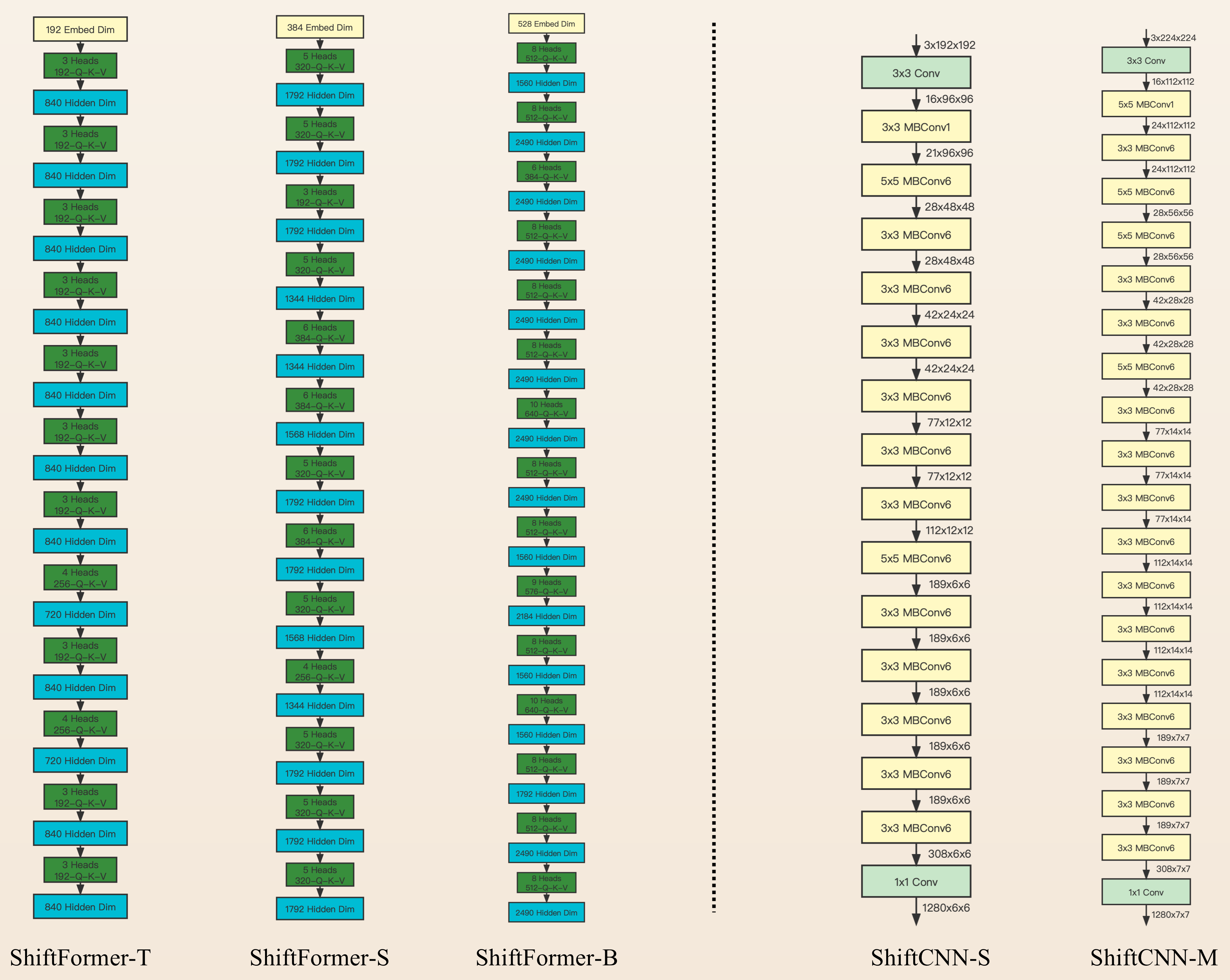}
    \caption{The searched architectures of ShiftNAS family models.}
    \label{fig:archs}
\end{figure*}

We show the searched architectures of ShiftNAS family models in \textbf{Figure \ref{fig:archs}}, including ShiftFormer-T, ShiftFormer-S, ShiftFormer-B, ShiftCNN-S and ShiftCNN-B.
\end{document}